\def\year{2022}\relax
\documentclass[letterpaper]{article} 
\usepackage{caption}    
\usepackage{subcaption} 
\usepackage{aaai22}  
\usepackage{times}  
\usepackage{helvet}  
\usepackage{courier}  
\usepackage[hyphens]{url}  
\usepackage{graphicx} 
\usepackage{natbib}  
\usepackage{caption} 
\usepackage{amsmath}
\usepackage{amsfonts}
\DeclareCaptionStyle{ruled}{labelfont=normalfont,labelsep=colon,strut=off} 
\usepackage{algorithm}
\usepackage{algorithmic}

\usepackage{newfloat}
\usepackage{listings}
\floatstyle{ruled}
\newfloat{listing}{tb}{lst}{}
\floatname{listing}{Listing}
\title{Visual Explanations for Convolutional Neural Networks\\ via Latent Traversal of Generative Adversarial Networks}
\author{
    Amil Dravid\textsuperscript{\rm 1}\thanks{With additional support from Florian Schiffers, Dr. Oliver Cossairt, Dr. Boqing Gong }, Aggelos K Katsaggelos\textsuperscript{\rm 2} \\
   
}
\affiliations{
    \textsuperscript{\rm 1} Northwestern University Computer Science, Evanston, IL 60208, USA \\
    \textsuperscript{\rm 2} Northwestern University ECE (Courtesy Computer Science and Radiology), Evanston, IL 60208, USA \\

    amildravid2023@u.northwestern.edu, a-katsaggelos@northwestern.edu
}

\usepackage{bibentry}

\begin{document}

\maketitle

\begin{abstract}
Lack of explainability in artificial intelligence, specifically deep neural networks, remains a bottleneck for implementing models in practice. Popular techniques such as Gradient-weighted Class Activation Mapping (Grad-CAM) provide a coarse map of salient features in an image, which rarely tells the whole story of what a convolutional neural network (CNN) learned. Using COVID-19 chest X-rays, we present a method for interpreting what a CNN has learned by utilizing Generative Adversarial Networks (GANs). Our GAN framework disentangles lung structure from COVID-19 features. Using this GAN, we can visualize the transition of a pair of COVID negative lungs in a chest radiograph to a COVID positive pair by interpolating in the latent space of the GAN, which provides fine-grained visualization of how the CNN responds to varying features within the lungs.     
\end{abstract}

\section{Introduction}
Interpreting CNNs has gained significant relevance with the surge of deep learning-enabled COVID detection models. However, many of these models have been found to be biased and misled by validation and visualization techniques such as Grad-CAM~\cite{degrave2021ai, selvaraju2017grad}. 

Generative Adversarial Networks (GANs) show promise for the task of feature visualization as they have gained considerable popularity in generating photo-realistic images~\cite{goodfellow2014generative}. A GAN consists of a discriminator and generator model that are trained in tandem. The generator learns to “fool” a discriminator model by trying to replicate the distribution of true data examples. It learns to map points from a low-dimensional manifold known as the \emph{latent space} via a vector of randomly sampled numbers. The generator transforms this vector into an image. The discriminator determines whether inputted data is “true” data from the actual distribution, or synthesized data from the generator. After training, it can be observed how one image morphs into another image by linearly interpolating between the two images' corresponding latent vectors. This provides the basis for our proposed method of feature visualization.  

\begin{figure}
    \centering
    \hspace*{-0.5cm} 
    \includegraphics[scale = 0.3]{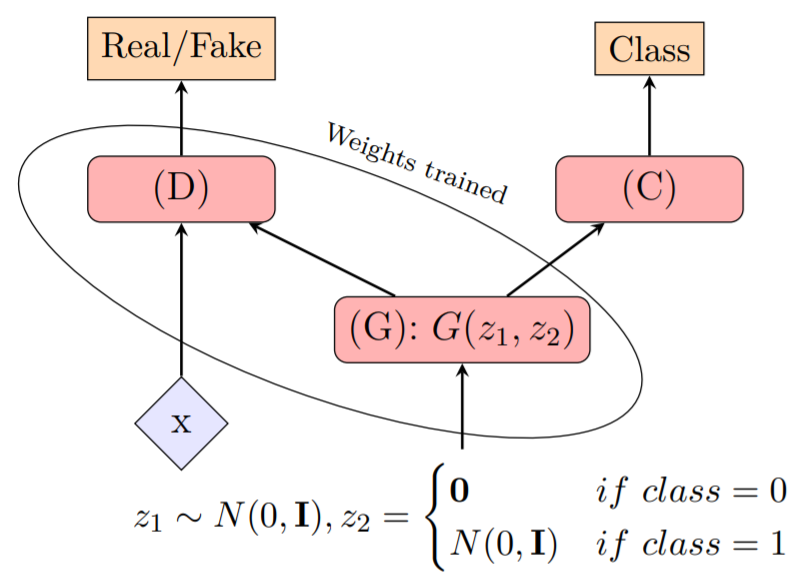}
    \caption{Generator (G) takes in structural latent vector $z_1$ and class latent vector $z_2$ to produce fake chest X-rays fed into the discriminator, along with real samples $x$. The classifier (C) provides feedback for generating class-discriminable images.}
    \label{fig:gan_diagram}
\end{figure}

\section{Methods}
Our method first relies on a pre-trained classifier that we wish to visualize. We specifically use a VGG16 model trained to $\sim$ 75\% accuracy on a private COVID chest X-ray dataset of 128x128 grayscale images. The GAN framework is inspired by the Auxiliary-Classifier GAN~\cite{odena2017conditional}, except we decouple the classifier from the discriminator, and employ a different latent vector scheme (see Figure \ref{fig:gan_diagram}). The generator carries out supervised disentanglement by taking in a latent vector $z_1$ that corresponds to lung structure, and a class information vector $z_2$. The vector $z_1$ is sampled from a spherical normal distribution. The $z_2$ sampling scheme relies on the intuition that COVID manifestations are not deterministic: the same pair of healthy lungs will retain their lung structure even with COVID, but COVID features can present in many ways within the lungs. Thus, when the class is COVID-negative (class = 0), $z_2$ is a vector of zeros, otherwise (class = 1) it is drawn from the spherical normal distribution to represent a continuous manifold of COVID features.

During training, the following objective is optimized:
\begin{equation}
\begin{aligned}
    \underset{G} {min} \text{ } \underset{D} {max} \text{ } &
    \underset{{\emph{x} \sim p_{x}}} {\mathbb{E}} [\log D(x)] +
    \underset {{z_1 \sim p_{z_1}, y \sim p_{y}}} {\mathbb{E}} [ \log (1- D(G(z_1, y )))]\\
    &-\underset {{z_1 \sim p_{z_1}, y \sim p_{y}}} {\mathbb{E}} [ \log ( p_c(y|G(z_1, y )))]
    \end{aligned}
\end{equation}

The first two terms correspond to the typical min-max game between the generator $G$ and discriminator $D$, where $x$ corresponds to data observations, $z_1$ is the structural latent vector, and $y$ is the class that is encoded in the $z_2$ vector. The third term relates to the generator learning to generate images that the classifier $C$ can correctly classify as COVID negative or positive. In this formulation, the generator is trained with discriminator to produce high-fidelity images, while getting feedback from the frozen classifier to incorporate class-specific features. It has been shown that minimizing this third term roughly approximates the KL divergence between the classifier's learned distribution classifier's $p_{c}(y|x)$ and the generator's $p_{g}(y|x)$~\cite{gong2019twin}. Thus, the generator provides a representation of what the classifier has learned. 

After training, the generator can be leveraged to explain the classifier. Given a COVID-positive image $x$, the latent vectors can be reconstructed by optimizing:

\begin{equation}
    \begin{aligned}
    \underset{z_1, z_2}{\text{arg min }}\text{MSE}(G(z_1,z_2),x) + \text{BCE}(C(G(z_1,z_2)), C(x))
    \end{aligned}
\end{equation}
The latent vectors $z_1$ and $z_2$ are found via gradient descent. The objective is to minimize the mean-squared error between the generated image and the ground-truth in addition to the binary cross-entropy between the classifier's output on both images. These two terms can be balanced with constant coefficients. After $z_1$ and $z_2$ are found, we can rely on the sampling scheme for $z_2$, changing it to $\mathbb{\overrightarrow {0}}$ to convert the COVID-positive lungs to COVID-negative. 

Finally, we can traverse the latent space to visualize how the classifier's output changes with the pathology within the lungs. We interpolate through the latent vector $z_2$ with steps $n$ at a rate of $\lambda$ while keeping the lung structure constant with $z_1$ by looking at the outputs of $G(z_1, \overrightarrow{0} + n\lambda z_2)$, for $n = 1, 2,...$

\section{Experiments and Results}
After training the generator for 1000 epochs, we evaluate how well $z_2$ maps to COVID features. We generate 4 samples from the same lung structure $z_1$, generating 1 COVID negative lung with $z_2 = \overrightarrow{0}$ and 3 positive with randomly drawn from $z_2 \sim N(0,\textbf{I})$. This is repeated 1000 times, and all samples are fed into the classifier. The classifier's predictions match the class fed into the generator with $91.15\%\pm{0.09}$ accuracy. Given that random guessing would yield $50\%$, the $z_2$ sampling scheme seems to incorporate COVID features as per the classifier.

\begin{figure}[t]
     \centering
     \begin{subfigure}[t]{0.5\textwidth}
         \centering
         \includegraphics[width=0.6\textwidth]{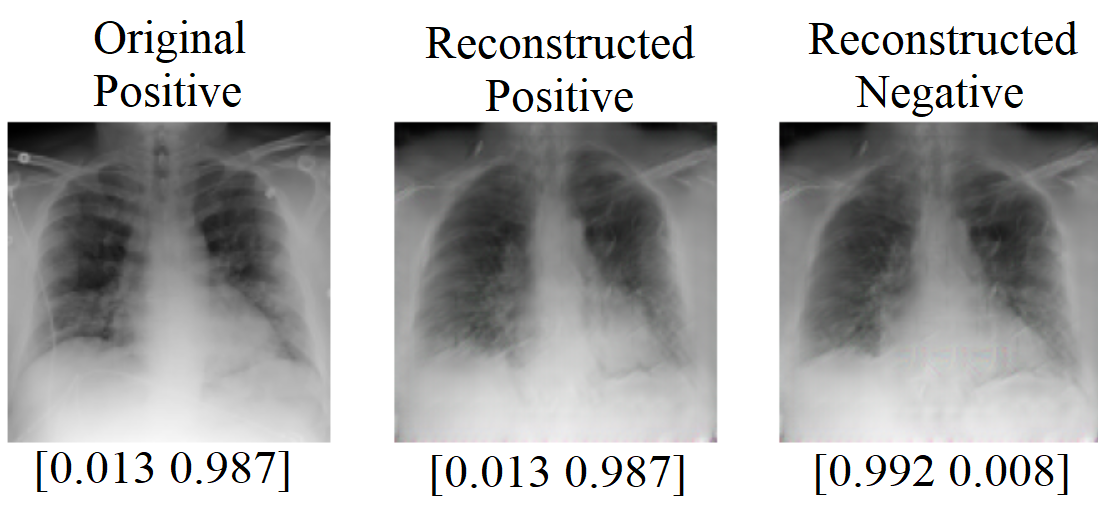}
         \caption{Reconstructing an original, real COVID positive image and turning it into negative with the generator. }
         \label{fig:reconstruct}
     \end{subfigure}
     \begin{subfigure}[t]{0.5\textwidth}
         \centering
         \includegraphics[width=0.775\textwidth]{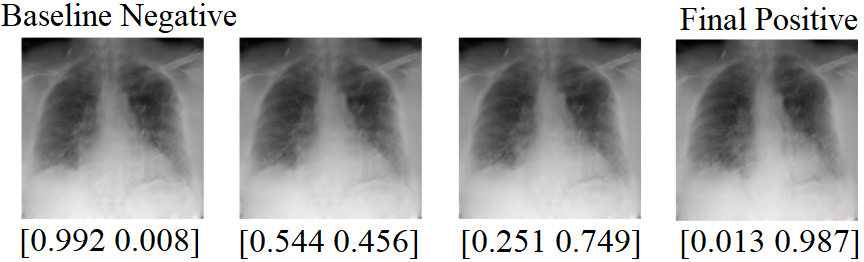}
         \caption{Interpolating through the latent space between the reconstructed pair of lungs with COVID and without.  }
         \label{fig:latent}
     \end{subfigure}
      \begin{subfigure}[t]{0.5\textwidth}
         \centering
         \includegraphics[width=0.55\textwidth]{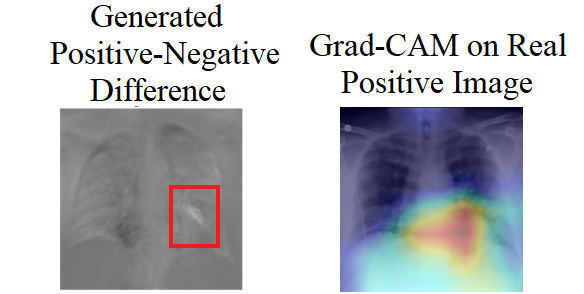}
         \caption{Pixel-wise difference between the last and first images in the latent interpolation highlights changing regions as the pair of lungs turns COVID positive. The most active region is highlighted. Compare against Grad-CAM to the right.}
         \label{fig:compare}
     \end{subfigure}
        \caption{Running through the proposed feature visualization pipeline. Classifier's softmax outputs below each example.}
        \label{fig:fig2} 
\end{figure}

When interpolating over the $z_2$ latent space between pairs of COVID negative and COVID positive lungs with the same $z_1$, the classifier's softmax probability for COVID positive monotonically increases as $z_2$ moves away from $\overrightarrow{0}$, which suggests that the $z_2$ latent space is structured such that $\overrightarrow{0}$ corresponds to the mean of a highly dense COVID negative probability region. This can be exploited in feature visualization. After reconstructing the COVID positive image and its negative pair with high confidence (as seen in Figure \ref{fig:reconstruct}), we can observe the softmax probabilities over the outputs as we morph the negative image into a positive (Figure \ref{fig:latent}). Thus, the images across the decision boundary can be observed as the classifier's prediction changes. Compared to Grad-CAM (Figure \ref{fig:compare}), traversing through the latent space provides more fine grained feature visualization and holds more explaining power. 

\bibliography{LaTeX/aaai22.bib}

\end{document}